\documentclass[letterpaper]{article} 
\usepackage{aaai23}  
\usepackage{times}  
\usepackage{helvet}  
\usepackage{courier}  
\usepackage[hyphens]{url}  
\usepackage{graphicx} 
\usepackage{natbib}  
\usepackage{caption} 
\usepackage{algorithm}
\usepackage{algorithmic}
\usepackage{adjustbox} 
\usepackage{booktabs}
\usepackage{diagbox}
\usepackage{multirow}
\usepackage{amsmath}
\usepackage{amsfonts}
\usepackage{amssymb}
\usepackage{makecell}
\usepackage{subfigure}
\usepackage{colortbl}
\usepackage{xcolor}
\usepackage[switch]{lineno}

\usepackage{newfloat}
\usepackage{listings}
\DeclareCaptionStyle{ruled}{labelfont=normalfont,labelsep=colon,strut=off} 
\floatstyle{ruled}
\newfloat{listing}{tb}{lst}{}
\floatname{listing}{Listing}
\title{Decorate the Newcomers:\\Visual Domain Prompt for Continual Test Time Adaptation}
\author{
    Yulu Gan\textsuperscript{\rm 1},
    Yan Bai\textsuperscript{\rm 1},
    Yihang Lou\textsuperscript{\rm 2},
    Xianzheng Ma\textsuperscript{\rm 3},
    Renrui Zhang\textsuperscript{\rm 4},
    Nian Shi\textsuperscript{\rm 5},
    Lin Luo\textsuperscript{\rm 1}\thanks{Corresponding author.}
}
\affiliations{
    \textsuperscript{\rm 1}Peking University, 
    \textsuperscript{\rm 2}Huawei Technologies,
    \textsuperscript{\rm 3}Wuhan University,
    \textsuperscript{\rm 4}The Chinese University of Hong Kong,\\
    \textsuperscript{\rm 5}Aerospace Information Research Institute, Chinese Academy of Sciences
    
ganyulu@stu.pku.edu.cn,\{yanbai,zhangrenrui,luol\}@pku.edu.cn, \\louyihang1@huawei.com, maxianzheng@whu.edu.cn, shinian.work@gmail.com
}

\usepackage{bibentry}

\begin{document}

\maketitle

\begin{abstract}
Continual Test-Time Adaptation (CTTA) aims to adapt the source model to continually changing unlabeled target domains without access to the source data. Existing methods mainly focus on model-based adaptation in a self-training manner, such as predicting pseudo labels for new domain datasets. Since pseudo labels are noisy and unreliable, these methods suffer from catastrophic forgetting and error accumulation when dealing with dynamic data distributions. Motivated by the prompt learning in NLP, in this paper, we propose to learn an image-level visual domain prompt for target domains while having the source model parameters frozen. During testing, the changing target datasets can be adapted to the source model by reformulating the input data with the learned visual prompts. Specifically, we devise two types of prompts, i.e., domains-specific prompts and domains-agnostic prompts, to extract current domain knowledge and maintain the domain-shared knowledge in the continual adaptation. Furthermore, we design a homeostasis-based prompt adaptation strategy to suppress domain-sensitive parameters in domain-invariant prompts to learn domain-shared knowledge more effectively. This transition from the model-dependent paradigm to the model-free one enables us to bypass the catastrophic forgetting and error accumulation problems. Experiments show that our proposed method achieves significant performance gains over state-of-the-art methods on four widely-used benchmarks, including CIFAR-10C, CIFAR-100C, ImageNet-C, and VLCS datasets.
\end{abstract}

\section{Introduction}

\begin{figure}[!ht]
\centering
\includegraphics[width=\linewidth]{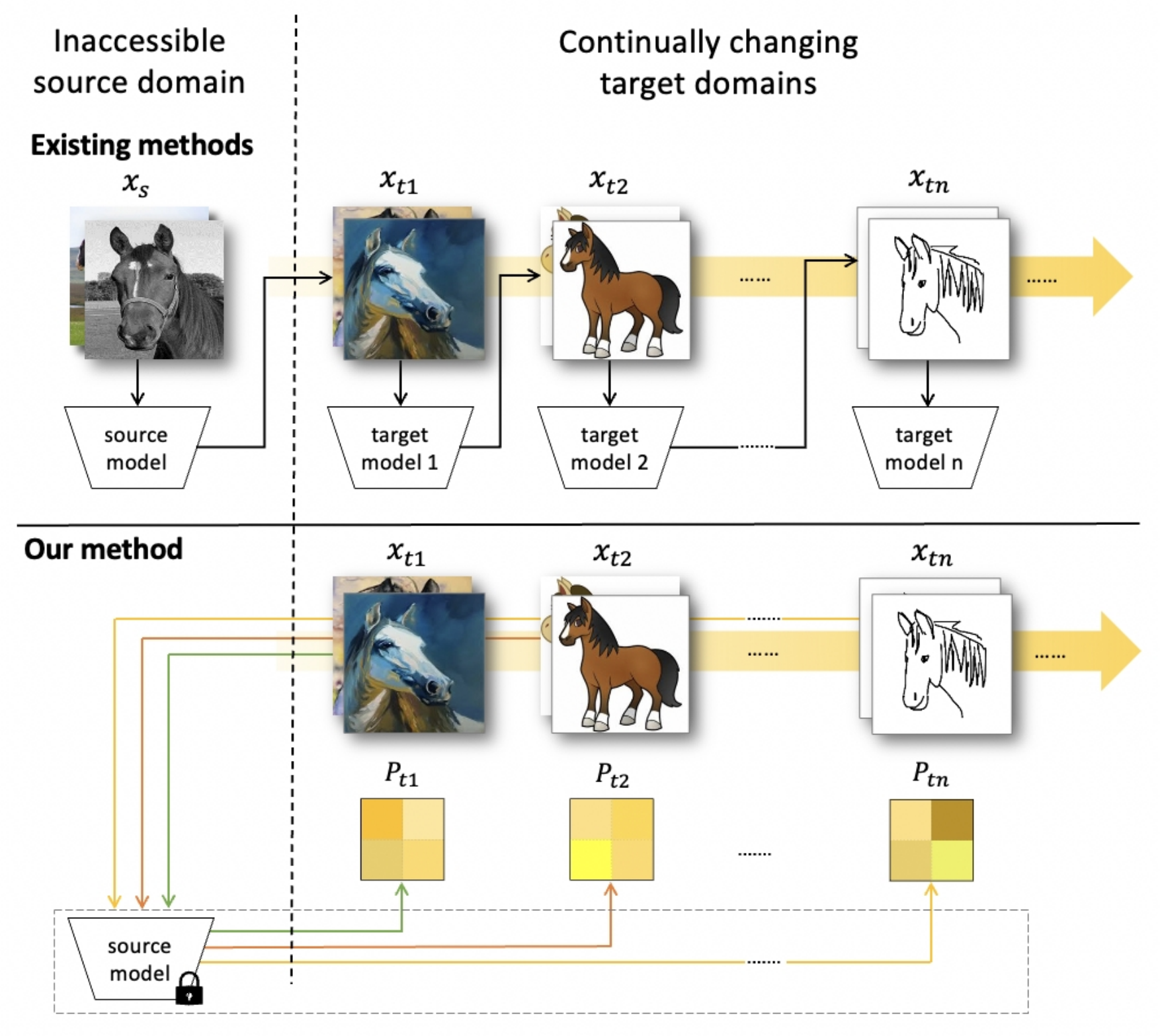}
\vspace{-0.5cm}
\caption{\textbf{The problem and our main idea.} Our goal is to adapt the source model to continually changing target domains. Existing methods focus on model-based adaptation. Differently, we design our method to adapt the changing target datasets to the source model by 1) tuning the visual prompts for each domain while keeping the source model frozen; 2) reformulating the input data with learned prompts when testing on the current target domain. This transition from the model-dependent paradigm to a model-free one enables us to eliminate catastrophic forgetting and achieve significant performance gains over state-of-the-art methods.}\label{intro}
\vspace{-0.5cm}
\end{figure}

\begin{table*}[htb]
\centering
\vspace{-0.2cm}
\caption{Adaptation settings differ by their data and corresponding losses during training and testing.}\label{compare settings} 
\setlength\tabcolsep{5pt}
\renewcommand\arraystretch{1.1}
\begin{tabular}{lcccccc}
\toprule
\textbf{setting} & \textbf{source data} & \textbf{target data} & \textbf{train loss} & \textbf{test loss} \\\midrule
fine-tuning & $ - $ & $x_t,y_t$ & $L(x_t,y_t)$ & $ - $ \\
domain adaptation & $x_s,y_s$ & $x_t$ & $L(x_s,y_s) + L(x_s,x_t)$ & $ - $ \\
test-time training & $x_s,y_s$ & $x_t$ & $L(x_t,y_t) + L(x_s)$ & $ - $ \\
test-time adaptation & $ - $ & $ x_t $ & $ - $ & $L(x_t)$ \\
continual test-time adaptation & $ - $ & $ x_{t1}\rightarrow...x_{tn}$ (continually changing) & $ - $ & $L(x_t)$ \\\bottomrule
\end{tabular}
\vspace{-0.2cm}
\end{table*}


The deep neural networks can perform well when the models are trained with supervision and the data follow a consistent distribution. However, the performance can drop significantly when a domain gap exists between the training and test data, especially when the distribution of the test data changes over time unexpectedly. Besides, this domain gap commonly exists in the real world \cite{Radosavovic2022}: for a model deployed on the street for signal light recognition, the target domain is constantly changing, such as foggy, snowy, rainy days, or the light difference between day and night. A source model needs to adapt to the heterogeneous and dynamic target domains to ensure acceptable performance in real applications. To this end, Continual Test-Time Adaptation (CTTA) is introduced and draws growing attention in the community.

Recently, researchers propose Test-Time Adaptation(TTA) to fine-tune the model or change the model's output distributions online during the test time. As shown in Table \ref{compare settings}, TTA cannot utilize source data during test time. Some works \cite{DequanWangetal2021, Liangetal2020, Chenetal2022} handle TTA by adjusting the network structures or fine-tuning several parts of the model. To avoid the degradation risk in modifying the source network \cite{Boudiafetal2022} realising TTA in a different perspective that only changes the model's outputs using the Laplace adaptive maximum likelihood estimation can be helpful. However, these TTA methods require the target domains to be of stationary distributions and thus neglect continuously changing target domains.

Given the environment is constantly changing over time, CoTTA \cite{Wangetal2022} first proposes to address a sequence of different domain shifts in CTTA, rather than a single shift in TTA. With the purpose of preventing error accumulation, CoTTA refines pseudo labels by the weight and data augmentation averages. To further avoid catastrophic forgetting, a small part of the neurons are randomly restored as the source pre-trained parameters during each iteration. CoTTA has considered the most essential problems of this task. 
However, being estimated from the source network, the pseudo labels are still unreliable and play a limited role in avoiding error accumulation, especially when the domain gap is large. Meanwhile, catastrophic forgetting cannot be fully solved by random restoration at the model level under a large domain gap. 

To tackle the error accumulation problem of CTTA, we first introduce a concept of visual domain prompts that are 1) small image tokens and 2) dynamically added upon the input images to shift them from the changing target domains to the source domain. We then propose a new CTTA framework (Figure~\ref{intro}), Visual Domain Prompt for Continual Test Time Adaptation, that consists of the visual domain prompt updating module and the Homeostasis-based adapting strategy. 
The visual domain prompt includes two types of tokens to adapt the new-coming images to the source model and to avoid over-fitting and catastrophic forgetting during adaptation. Meanwhile, a homeostasis-based adapting strategy regularizes the domain-sensitive parameters in the prompt to prevent the model from having a strong bias against the data of the current domain.
Different from the previous work where the model fine-tuning sometimes leads to model degradation~\cite{Boudiafetal2022}, we solve this problem from the perspective of changing the input images by the light-weight domain prompt tokens, achieving good performance at a relatively small cost.

Our main contributions are highlighted as follows:
\begin{itemize}
\item To the best of our knowledge, we propose the first lightweight prompt approach that handles the CTTA problem from the input image level. By using the visual domain prompts to dynamically update a small portion of the input image pixels, domain adaption is achieved while the error accumulation problem is mitigated.
\item We further introduce a Homeostasis-based prompt adapting strategy to avoid catastrophic forgetting by limiting domain-sensitive parameters from over-adaptation. 
\item  Our proposed approach outperforms most state-of-the-art methods according to the experiments on extensive benchmark datasets, covering both synthetic and real-world domain gaps. It proves that our approach is practical for both good performance and low cost.
\end{itemize}

\section{Related Work}

\begin{figure*}[htb]
\centering
\includegraphics[width=\linewidth]{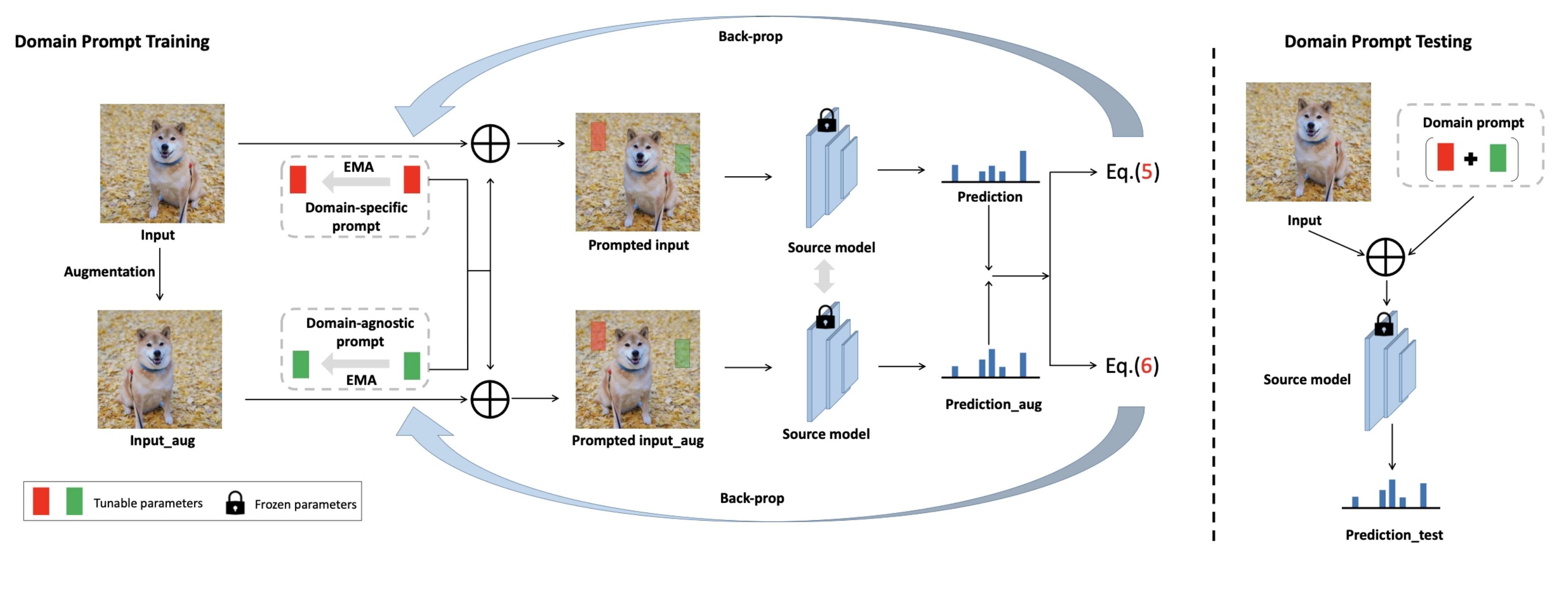}
\vspace{-0.3cm}
\caption{\textbf{The whole framework.} (a) \textbf{Domain prompt training}. As we freeze the source model, all update processes only work on the domain-specific prompt and the domain-agnostic prompt. We use the teacher-student framework to update these two prompts. The teacher network stops back-propagation, and we use the student network to learn. When entering each training time step t, the student network will transfer its parameters to the teacher network by exponential moving average (EMA) updating. The model's inputs are the original images and images with augmentation. The domain-specific prompt and domain-agnostic prompt are summed point by point to obtain the prompted input and the prompted input with augmentation, respectively. The prediction results are obtained through the frozen source model. We used equation (\ref{eq2}) to update the domain-specific prompt, and utilize equation (\ref{eq3}) to optimize the domain agnostic prompt. Because the update process is performed at the batch level, the prompts learned from the previous batch level will be applied to the image of the next batch. (b) \textbf{Domain prompt testing}. During the domain prompt testing time, the input image is summed up with the domain-specific and domain-agnostic prompts. Then we feed this reformulated image into the source domain model for prediction.}\label{fig:2} 
\vspace{-0.3cm}
\end{figure*}

\subsection{Test-time adaptation}
Test-time adaptation (TTA) aims to fine-tune the source model only by the upcoming data in the test time. TTA cannot rely on source data or supervision. \cite{DequanWangetal2021}, and mainly divided into two categories. One is the network-based method: Similar to the adaptive batch-norm methods \cite{Burnsetal2021,Lietal2018,Nadoetal2020,Schneideretal2020}, Tent\cite{ZifengWangetal2021} fine-tune the batch normalization parameters by minimizing entropy loss. Similarly, Shot\cite{Liangetal2020} updates the parameters of the feature extractor of a given model by maximizing the mutual information loss. The method proposed in \cite{Chenetal2022} combines contrastive learning and self-training to solve this problem. The other is the parameters-free method. In \cite{Boudiafetal2022}, instead of updating the model's parameters, they adjust the output distribution by Laplace maximum likelihood.

It's worth noting that CoTTA \cite{Wangetal2022} is the first work to consider continual distribution shifts in the real world and provide a model-based approach. However, we try to solve this problem from the input level.

\subsection{Prompt learning}
Prompt learning is first proposed in natural language processing (NLP), hoping to adapt pre-trained visual-language models to various downstream tasks. Recently, the idea of prompt has been transferred to some multi-modal tasks \cite{ShengShen2022MultitaskVP}. CoOp \cite{zhouetal2022coop} introduces prompt learning into the computer vision to adapt to the pre-trained visual-language model by transforming the context word into a set of learnable vectors. CoCoOp \cite{zhouetal2022cocoop} further improved this static prompt to a dynamic prompt to better adapt to class shift. However, they do not design for domain shift. In \cite{Lietal2021}, a new pre-training task is proposed, which learns the fine-grained alignment between visual regions and text entities in a self-supervised manner through the entity prompt module. \cite{Geetal2022} is the first work to utilize prompt learning in domain adaptation. In addition to adding prompts in the text modal, some work tries other prompt forms. \cite{Bahngetal2022,MenglinJia2022VisualPT} proposes a new representation of visual prompts for pre-trained transformers. \cite{ZhiboWangetal2021} takes the features as prompts and dynamically learns the prompt to guide the model.

Our method is similar to prompt learning. However, unlike the previous works, our proposed method is text-modal-free. Moreover, we give additional thought to the problem of a sequence of domain shifts. Our method proposes a token-guided conduction module to adapt to current data quickly. Apart from only tuning the token-guided conduction module, we apply the Homeostatic-plasticity-based adapting strategy to restrain this module, effectively against catastrophic forgetting in continual adaptation scenarios.

\subsection{Continual learning} 
Continual learning wishes the model to aggregate knowledge in a new domain without forgetting the previous knowledge in old domains. For this purpose, three kinds of methods are proposed. The first is the replay method. A reservoir sampling method is proposed in \cite{davidrolnicketal2018} to limit the number of samples stored. The second is the regularization-based methods, \cite{octomiaoetal2016, FriedemannZenkeetal2017} proposes the idea of elastic weight estimation and adjusts the weight updating strategy according to the importance estimate of the parameters. The third type is the parameter isolation method, which provides different model parameters for each task to prevent any possible forgetting~\cite{matthiasdelangeetal2021}.

The aforementioned methods address the setup of continually learning new tasks or classes. In contrast, the problem we focus on aims to solve the catastrophic forgetting problem when facing different domains with the same task. 

\section{Method}
Given a model $q_\theta(y|\boldsymbol{x})$ trained on the source domain and to be tested on multiple target datasets $ {\widetilde D_{\mu1}, \widetilde D_{\mu2} , ... , \widetilde D_{\mu T}}$, where $\widetilde D_{\mu i}={\{(x_i^T)\}}_{i=1}^{N_\mu}$, and $N_\mu$ represents the scale of the target domain, the distributions of the target domains can change or reoccur over time (i.e., sunny, cloudy, rainy, and night domains in the scene of automatic driving). Our goal is to continuously adjust the incoming data of new distributions online to adapt to the source model.  

Our approach freezes the source model to avoid error accumulation and catastrophic forgetting issues of model-tuning methods, and only modifies the arriving input images. The framework includes two modules, a visual domain prompt module (including two types of prompts) and a homeostatic-based prompt updating strategy (Figure \ref{fig:2}). 

\subsection{Continual test time adaptation via domain prompt}
The proposed lightweight visual domain prompts include two types: domain-specific vs, domain-agnostic prompts.  
As a messenger between the input data and the models, the visual domain prompts are directly added upon a portion of the input image and the decorated image is then input into the source model for predictions. The prompts are updated periodically following two different loss functions.

\noindent\textbf{Design of the two different prompts.} 
Inspired by text prompt learning in natural language processing(NLP) which guides the model to correct predictions \cite{PengfeiLiu2021PretrainPA}, we propose two visual prompts with different functions for the continual test time adaptation task. One is a domain-specific prompt (DSP), and the other is a domain-agnostic prompt (DAP).
The domain-specific prompt aims to extract current domain knowledge, while the domain-agnostic prompt produces an effect on maintaining the domain-shared knowledge.
We define DSP as $\psi_\delta$ and $\omega_\phi$ for DAP. equation \ref{eq1} shows that DSP and DAP are both learnable 
parameter matrices that apply to the input images $x^T$ by 
summing up. $x^T$, $\omega_\phi$ and $\psi_\delta$ are summed up point by point according to the coordinates. Regions without overlap do not be operated. We then obtain the reformulated image $x^{T}_{p}$ for training and testing. Besides, It is worth noting that the size and position of DSP and DAP are flexible.

\begin{equation} \label{eq1}
\begin{aligned}
x^{T}_{p} = x^T + \omega_\phi + \psi_\delta.
\end{aligned}
\end{equation}


Hebbian theory \cite{DOHebbetal1988} shows that Plasticity is a synaptic mechanism which detects and amplifies co-activity between neurons. This mechanism includes Hebbian Plasticity and homeostatic Plasticity, aiming to stabilize neuronal activity in a short time rapidly. \cite{KennethDMilleretal1994,SenSongetal2000, FriedemannZenkeetal2017}.

We get inspiration from the Hebbian theory to achieve the functions of DSP and DAP. We apply different loss functions to DSP and DAP to ensure the plasticity of neurons and enhance their stability. More details are in the Homeostatic-based adapting strategy for updating prompts.  

\noindent\textbf{Updating mechanism for prompts.}
We use the teacher-student network as the framework to update the two prompts. When the student network ${f_\theta}_t$ updates from $\theta_t \to \theta_{t+1}$, the weights of the student network will be updated to the teacher network ${f_\theta'}_t$ by exponential moving average updating. Note that updating the parameters is only performed on the two prompts, as the source model is frozen. We utilize cross-entropy loss as the optimization function of DSP. For DAP, we add a regularization term to constrain those domain-sensitive parameters to alleviate the instability when facing domain changes. We will apply the learned prompts from the previous batch to the next batch for testing. 


\begin{table*}[htb]
\caption{\label{CIFAR100-to-CIFAR100c} Classification error rate (\%) for the standard CIFAR100-to-CIFAR100C online continual test-time adaptation task. All results are evaluated on the ResNeXt-29 architecture with the largest corruption severity level 5. Our method far exceeds the state-of-the-art methods by 16.2\%. Gain(\%) represents the percentage of improvement in model accuracy compared with the source method.}
\centering
\setlength\tabcolsep{2pt}
\begin{adjustbox}{width=1\linewidth,center=\linewidth}
\begin{tabular}{lccccccccccccccccc}
\toprule
Method & gaussion & shot & impulse & defocus & glass & motion & zoom & snow & frost & fog & brightness & contrast & elastic\_trans & pixelate & jpeg & Mean$\downarrow$ & Gain\\\midrule
Source &73.0&68.0&39.4&29.3&54.1&30.8&28.8&39.5&45.8&50.3&29.5&55.1&37.2&74.7&41.2&46.4&0.0\\
BN Stats Adapt \cite{Schneideretal2020} &42.1&40.7&42.7&27.6&41.9&29.7&27.9&34.9&35.0&41.5&26.5&30.3&35.7&32.9&41.2&35.4&+11.0\\
Pseudo-label \cite{Leeetal2013} &38.1&36.1&40.7&33.2&45.9&38.3&36.4&44.0&45.6&52.8&45.2&53.5&60.1&58.1&64.5&46.2&+0.2\\
Tent-continual \cite{DequanWangetal2021}  &37.2&35.8&41.7&37.9&51.2&48.3&48.5&58.4&63.7&71.1&70.4&82.3&88.0&88.5&90.4&60.9&-14.5\\
CoTTA \cite{Wangetal2022} &40.1&37.7&39.7&26.9&38.0&27.9&26.4&32.8&31.8&40.3&24.7&26.9&32.5&28.3&33.5&32.5&+13.9\\
\cellcolor{lightgray}\textbf{Ours (proposed)} &\cellcolor{lightgray}\textbf{29.5}&\cellcolor{lightgray}\textbf{25.8}&\cellcolor{lightgray}\textbf{31.9}&\cellcolor{lightgray}\textbf{2.8}&\cellcolor{lightgray}\textbf{30.5}&\cellcolor{lightgray}\textbf{7.7}&\cellcolor{lightgray}\textbf{5.7}&\cellcolor{lightgray}\textbf{14.8}&\cellcolor{lightgray}\textbf{14.8}&\cellcolor{lightgray}\textbf{24.2}&\cellcolor{lightgray}\textbf{1.8}&\cellcolor{lightgray}\textbf{6.8}&\cellcolor{lightgray}\textbf{18.5}&\cellcolor{lightgray}\textbf{9.1}&\cellcolor{lightgray}\textbf{28.0}&\cellcolor{lightgray}\textbf{16.8}&\cellcolor{lightgray}\textbf{+29.6} \\\bottomrule
\end{tabular}
\end{adjustbox}
\vspace{-0.3cm}
\end{table*}

\begin{table*}[htb]
\caption{\label{CIFAR10-to-CIFAR10C}Classification error rate(\%) for standard CIFAR10-to-CIAFAR10C online continual test-time adaptation task. Results are evaluated on WideResNet-28 with the largest corruption severity level 5. Our method exceeds the state-of-the-art methods by 2.3\%. Gain(\%) represents the percentage of improvement in model accuracy compared with the source method.}
\centering
\setlength\tabcolsep{2pt}
\begin{adjustbox}{width=1\linewidth,center=\linewidth}
\begin{tabular}{lccccccccccccccccc}
\toprule
Method & gaussion & shot & impulse & defocus & glass & motion & zoom & snow & frost & fog & brightness & contrast & elastic\_trans & pixelate & jpeg & Mean$\downarrow$ & Gain\\\midrule
Source&72.3&65.7&72.9&46.9&54.3&34.8&42.0&25.1&41.3&26.0&9.3&46.7&26.6&58.5&30.3&43.5&0.0\\
BN Stats Adapt \cite{Schneideretal2020} &28.1&26.1&36.3&12.8&35.3&14.2&12.1&17.3&17.4&15.3&8.4&12.6&23.8&19.7&27.3&20.4&+23.1\\
Pseudo-label \cite{Leeetal2013} &26.7&22.1&32.0&13.8&32.2&15.3&12.7&17.3&17.3&16.5&10.1&13.4&22.4&18.9&25.9&19.8&+23.7\\
Tent-online \cite{DequanWangetal2021}  &24.8&23.5&33.0&12.0&31.8&13.7&10.8&15.9&16.2&13.7&7.9&12.1&22.0&17.3&24.2&18.6&+24.9\\
Tent-continual \cite{DequanWangetal2021} &24.8&20.6&28.6&14.4&31.1&16.5&14.1&19.1&18.6&18.6&12.2&20.3&25.7&20.8&24.9&20.7&+22.8\\
Baseline(CoTTA) \cite{Wangetal2022} &24.3&21.3&\textbf{26.6}&11.6&\textbf{27.6}&12.2&10.3&14.8&14.1&12.4&7.5&10.6&\textbf{18.3}&13.4&\textbf{17.3}&16.2&+27.3\\
\cellcolor{lightgray}\textbf{Ours (proposed)} &\cellcolor{lightgray}\textbf{22.6}&\cellcolor{lightgray}\textbf{19.7}&\cellcolor{lightgray}28.1&\cellcolor{lightgray}\textbf{7.1}&\cellcolor{lightgray}28.4&\cellcolor{lightgray}\textbf{9.5}&\cellcolor{lightgray}\textbf{6.3}&\cellcolor{lightgray}\textbf{10.2}&\cellcolor{lightgray}\textbf{11.5}&\cellcolor{lightgray}\textbf{9.0}&\cellcolor{lightgray}\textbf{1.5}&\cellcolor{lightgray}\textbf{5.6}&\cellcolor{lightgray}18.5&\cellcolor{lightgray}\textbf{12.8}&\cellcolor{lightgray}18.5&\cellcolor{lightgray}\textbf{13.9}&\cellcolor{lightgray}+\textbf{29.6}\\\bottomrule
\end{tabular}
\end{adjustbox}
\vspace{-0.3cm}
\end{table*}

\subsection{Homeostasis-based prompt adapting strategy}
This section will introduce the strategy to update the two prompts. We apply the cross-entropy loss (see equation \ref{eq2}) to optimize the DSP to learn domain-specific knowledge. As for DAP, to extract more domain-agnostic knowledge, we utilize an extra Homeostatic regularization. We reach the goal of mining the domain-agnostic knowledge by limiting parameters sensitive to domain changes. We will then introduce how we design this regularization term to make DAP learn domain-independent knowledge.

\noindent\textbf{Measure parameters' sensitivity toward domain shift.} We first evaluate the sensitivity of parameters to different domains. $g(\theta(t))=\frac{\partial \mathcal{L}}{\partial \theta(t)} $ is defined as the gradient at time $t$. $t_0$ and $t_1$ respectively denote a certain time in two adjacent target domains. As equation \ref{eq5} shown, we express the difference of losses as the product of gradient $g(\theta(t))$ and $\theta(t)'$. We use weight importance $\eta_i^\tau$ in Eq. \ref{eq5} to quantify the contribution of each parameter's value to the total loss.

\begin{equation} \label{eq5}
\begin{aligned}
\mathcal{L}(\theta_{t_1}) - \mathcal{L}(\theta_{t_0}) &=
\int_{t0}^{t1}g(\theta(t))d\theta \\&=
\int_{t0}^{t1}g(\theta(t)) \cdot \theta'(t)dt \\&=
- \sum\limits_i \eta_i^\nu,
\end{aligned}
\end{equation}

Given the weight importance $\eta_i^\nu$ and the difference $\delta_i^\nu=\theta_{i(t)}^\nu-\theta_{i(t-1)}^\nu$ between parameters of the domain $\nu$, the Homeostatic Factor is set as $\Lambda_i^\tau$:

\begin{equation}
\begin{aligned}
\Lambda_i^\tau = \sum\limits_{v<\tau}\frac{\eta_i^\nu}{(\delta_i^\nu)^2+\xi},
\end{aligned}
\end{equation}

Note that $\tau$ refers to the current domain and v denotes each domain before $\tau$.To get rid of the situation where $\delta_i^\nu \to 0$, we introduce $\xi=0.01$ into the Homeostatic Factor. $\delta_k^\nu$ being small means the parameter is changing slightly. And the parameters' contribution is more significant to the total loss when $\eta_k^\nu$ is larger. Therefore, when the $\delta_k^\nu$ is relatively small, or $\eta_k^\nu$ is relatively large, we believe the parameters $\theta_i$ are sensitive to the domain change. In equation \ref{penalty term}, this regularization term will penalize those parameters sensitive to domain shift. And update domain-insensitive parameters stably to consolidate the domain-agnostic knowledge. 

\begin{equation} \label{penalty term}
\begin{aligned}
{\mathcal{L}(\psi_\delta)} = \alpha \sum\limits_{\theta \in \Theta} \Lambda_i^\tau ||\theta - \theta^*||_2^2,
\end{aligned}
\end{equation}

We use $\alpha$ to control the contribution of the regularization term. When $\alpha$ takes a small value, the suppression of domain-sensitive parameters will be reduced. We define $\theta^*$ as the model's parameters of the last mini-batch of the previous domain. $\theta$ is the parameter that needs to be inferred. 

Note that $\Lambda_i^\tau$ will be updated when entering a new target domain to establish the connection between the previous and current domains. Once a new target domain comes, $\Lambda_i^\tau$ will be updated, and the $\eta_i^\nu$ is set to zero. Details for detecting the change of target domains are in the following subsection.

\noindent\textbf{Domain-shift detection.} To update the $\Lambda_i^\tau$, we need to know when the domain changes. Based on the observation of the prediction confidence $Conf_{(t)}$ of pseudo labels changing dramatically with the domain changing, we use it to estimate whether the target domain changes. Therefore, we set a threshold $S=0.25$. The difference between the prediction confidence will be continually calculated at the batch level. Once the $\Delta Conf=Conf_{(t+1)} - Conf_{(t)}$ is higher than the threshold $S$, $\theta^*$ and the $\eta_i^\nu$ will be updated.

\noindent\textbf{Overall loss for DSP and DAP.}
Finally, the loss functions for DSP and DAP can be equation \ref{eq2} and \ref{eq3}, respectively. Note that $h(x_p^T)$ indicate the random augment image.

\begin{equation} \label{eq2}
\begin{aligned}
\mathcal{L}_{\omega_\phi}(x_p^T)=-\sum\limits_C {f_\theta}_t'(h(x_p^T)) ({logf_\theta}_t(x_p^T)),
\end{aligned}
\end{equation}

\begin{equation} \label{eq3}
\begin{aligned}
\mathcal{L}_{\psi_\delta}(x_p^T)=-\sum\limits_C {f_\theta}_t'(h(x_p^T)) ({logf_\theta}_t(x_p^T))+\mathcal{L}(\psi_\delta),
\end{aligned}
\end{equation}

\begin{equation} \label{overall loss}
\begin{aligned}
\mathcal{L}_o(x_p^T)= \mathcal{L}_{\psi_\delta}(x_p^T) + \mathcal{L}_{\omega_\phi}(x_p^T),
\end{aligned}
\end{equation}

We use equation \ref{overall loss} to optimize DSP and DAP, to achieve the purpose of mining domain-specific knowledge and maintaining domain-agnostic knowledge.

\section{Experiments}

\begin{table*}[htb]
\caption{\label{dg benchmark} \textbf{Anti-forgetting performance on VLCS.} Classification error rate(\%) is used as the evaluate metric. VLCS has four domains. We take one as the source domain and the other three as the target domain. In the test time, the three target domains are sequentially tested for multiple rounds. Gain means the improvement of our method compared with CoTTA. Our method does not have the problem of catastrophic forgetting when encountering the same target domain at different rounds. Moreover, the models' performance is gradually improved, which shows the effectiveness of our proposed visual domain prompts.}
\centering
\setlength\tabcolsep{3pt}
\begin{adjustbox}{width=1\linewidth,center=\linewidth}
\begin{tabular}{c|c|ccccc|ccccc|ccccc|c|c }
\hline
\multicolumn{2}{c|}{Time}     & \multicolumn{15}{c}{$t$ \makebox[10cm]{\rightarrowfill} }                                                                              \\ \hline
\multicolumn{2}{c|}{Round}          & \multicolumn{5}{c|}{1}    & \multicolumn{5}{c|}{2}     & \multicolumn{5}{c|}{3}  & \multirow{2}{*}{Mean$\downarrow$}   & \multirow{2}{*}{Gain}  \\ \cline{1-17}
Method & \diagbox{Source}{Target} & Caltech-101 & VOC2007 & LabelMe & SUN09 & Mean$\downarrow$ & Caltech-101 & VOC2007 & LabelMe & SUN09  & Mean$\downarrow$ & Caltech-101 & VOC2007 & LabelMe & SUN09 & Mean$\downarrow$ & \\ \hline
\multirow{4}{*}{CoTTA} & Caltech-101 &35.4&61.2&65.7&61.5&56.0&35.7&60.6&66.2& 	61.9&56.1&35.9&61.6&67.2&64.1&57.2&56.4&/\\ 
 & VOC2007&32.6&64.9&75.7&37.3&52.6&32.6&70.0&82.8&43.9&57.3&35.0&76.2&86.9&52.6 &62.7&57.5&/  \\ 
 & SUN09 &46.0&51.2&8.1&57.9&40.8&44.4&50.3&7.9&57.1&39.9&43.6&49.7&8.4&57.0 	&39.7&40.1&/  \\  \hline
\multirow{4}{*}{\textbf{Ours}} & Caltech-101
&29.4&60.5&67.4&62.8&55.0&31.0&59.5&66.4&61.6&54.6&32.0&58.9&66.1&61.0&54.5&54.7&+1.7 \\ 
 & VOC2007 &29.8&46.5&53.5&33.0&40.7&28.8&46.0&55.1&36.2&41.5&28.8&46.9 	&55.9&38.8&42.6&41.6&+15.9 \\ 
 & SUN09 &48.4&46.1&7.0&55.0&39.1&48.3&46.0&6.9&54.1&38.8&47.9&45.8&6.9&52.6&38.&38.7&+1.4  \\  \hline
\end{tabular}
\end{adjustbox}
\vspace{-0.3cm}
\end{table*}

\subsection{Setup}
\textbf{Dataset} We evaluate our proposed method on five continual test-time adaptation and domain generalization benchmark tasks: CIFAR10-to-CIFAR10C(standard and gradual), CIFAR100-to-CIFAR100C and ImageNet-to-ImageNet-C. Moreover, to explore the ability to deal with the actual domain gap, we also evaluate our method on VLCS.

\noindent\textbf{Task setting} We follow CoTTA \cite{Wangetal2022} to set our four tasks, and we design a VLCS task to validate the anti-forgetting ability when deal with the real-world domain gap. For CIFAR10-to-CIFAR10C standard task, CIFAR100-to-CIFAR100C task and ImagetNet-to-ImagetNet-C task, given the source model, we need to adapt to the fifteen target domains with different corruption types that arrive sequentially. We evaluate all models under the largest corruption severity level 5. The evaluation is based on the online prediction results immediately after the data encounter. For the CIFAR10-to-CIFAR10C gradual task, the model will adapt to 0-5 corruption levels of a specific corruption type. Specifically, the corruption level will start from level 1 to level 5 and then gradually down to level 1. The exact process will be carried out on 15 corruption types in turn. For the VLCS task we designed, the VLCS dataset contains four domains. We use one of them as the source domain and the rest as the target domain for multiple rounds to validate the model's anti-forgetting ability.

\noindent\textbf{Implementation Details} In this paper, all experiments are conducted with PyTorch. We follow CoTTA \cite{Wangetal2022} to set our tasks. For CIFAR10-to-CIFAR10C standard and gradual tasks, CIFAR100-to-CIFAR100C tasks, the image size is 32 $\times$ 32, and the batch size is set to 100. Following the official public implementation from Tent \cite{DequanWangetal2021} , we adopt WideResNet-28 \cite{SergeyZagoruykoetal2016} model from the RobustBench benchmark \cite{FrancescoCroceetal2020} for CIFAR10-to-CIFAR10C standard and gradual tasks. And we adopt ResNeXt-29 \cite{ShiqiYangetal2021} model from \cite{DanHendrycksetal2019} for the CIFAR100-to-CIFAR100C task, which is used as one of the default architectures for CIFAR100 in the RobustBench benchmark \cite{FrancescoCroceetal2020}. The ImageNet-to-ImageNet-C \cite{DanHendrycksetal2019} experiments use the standard pre-trained Resnet50 model in RobustBench \cite{FrancescoCroceetal2020}. We train the source model for three epochs on the source domain to initialize our visual domain prompts.

\begin{figure}[!tb]
\centering
\includegraphics[width=\linewidth]{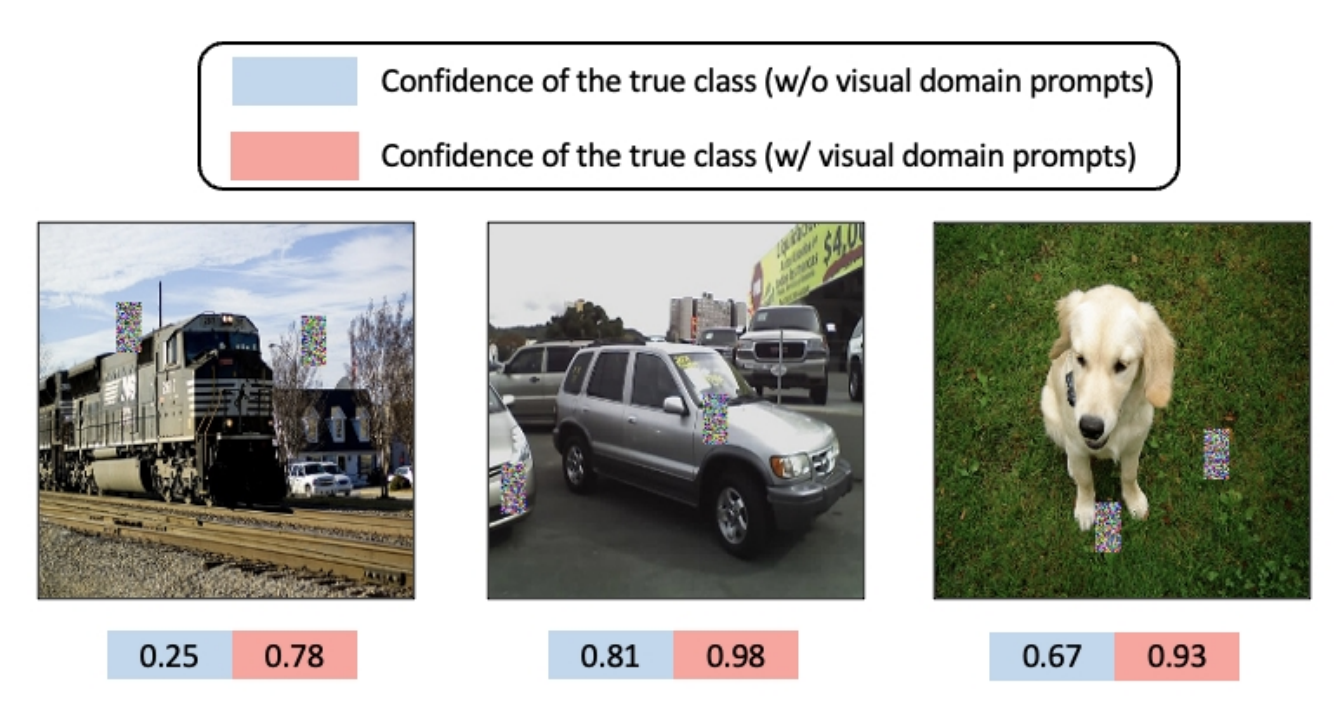}
\vspace{-0.5cm}
\caption{\label{fig:3}\textbf{Prediciton confidence from VOC in VLCS at the first round (Source model trained with the Caltech) and visualization of the visual domain prompts} Red/Blue represent predict confidence of the ground truth class with/without visual domain prompts. Predictions by applying the visual domain prompts show higher confidence.}
\label{visualization}
\vspace{-0.3cm}
\end{figure}

\subsection{Performance on synthesized domain shifts}
We evaluate our proposed method on four benchmark continual test time adaptation datasets with corruption type domain shift, i.e.,  CIFAR10-to-CIFAR10C stand tasks, CIFAR10-to-CIFAR10C gradual tasks, CIFAR100-to-CIFAR100C tasks, ImageNet-to-ImageNet-C.

\noindent\textbf{CIFAR10-to-CIFAR10C standard task.} Given source model trained on CIFAR10, we conduct TTA on CIFAR10C. There are fifteen corruption types that will sequentially come during the test time. As shown in Table \ref{CIFAR10-to-CIFAR10C},  the average error directly using the source domain model is up to 43.5\%. Recent advanced methods can reduce this error to 16.2\%. we can further reduce the error by 2.3\%. In our method, the errors of several corruption types, such as defocus, motion, zoom, fog, brightness, and contrast, are all below 10\%. Most of these belong to the corruption type of blur and weather type. We believe that weather-type corruption is a kind of additive corruption. Factors such as fog and rain are added to the original image. Our proposed method can learn more appropriate prompts according to the current data to guide the model to use the pre-training knowledge effectively.

\noindent\textbf{CIFAR10-to-CIFAR10C gradual task.} This task makes continuous changes among 0-5 levels of 15 corruption types. Table \ref{CIFAR10-to-CIFAR10C gradual} shows our results obtain the lowest error 6.0\%, 4\% improvement over the SOTA CoTTA method.

\noindent\textbf{CIFAR100-to-CIFAR100C}
CIFAR100-to-CIFAR100C is a more difficult task because it contains more categories than CIFAR10C. Surprisingly, as shown in Table \ref{CIFAR100-to-CIFAR100c}, our method performs far better on CIFAR100C than other existing methods. It still performs well on several blur types. 
In addition, it also performs better on two digital type corruptions: pixel and contrast.
We think this is because our proposed method learns domain-specific knowledge by adding tokens. This can guide the model to use pre-training knowledge.
The model can still perform well when the last few corruption types arrive, demonstrating our method's effectiveness.



\begin{figure*}[!htb]
\centering
\includegraphics[width=0.8\linewidth]{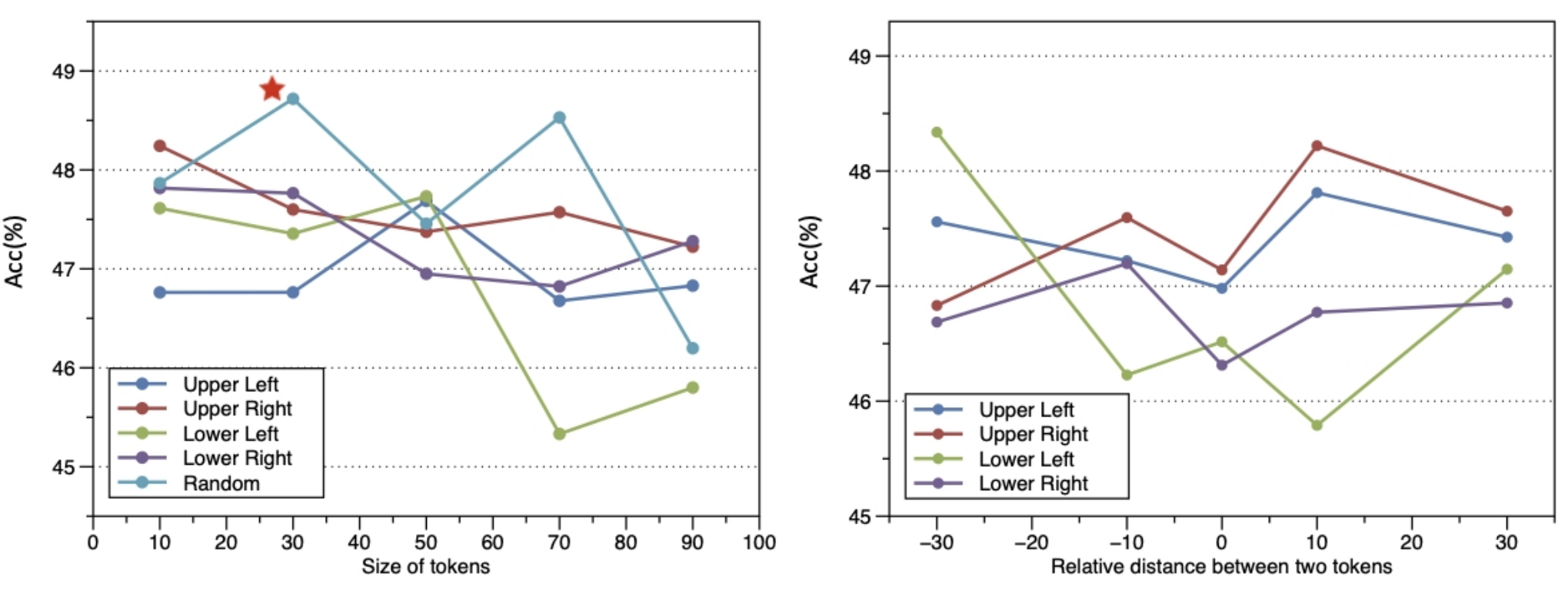}
\caption{\label{fig:4} \textbf{Effects of the prompts' size, location on the image and relative distance between the two prompts.} Experiments are conducted on the ImageNet-to-ImageNet-C task. The left figure shows the effect of the size and position of prompts on the model performance. When the prompts' size equals 30, and apply to the image randomly, the model's performance achieves the best. The right figure shows the effect of the relative positions. We set the prompt size to 20$\times$ 20. If the distance is negative,  the domain-independent prompt will be on the left side of the domain-specific prompt and vice versa. We find that exchanging the positions of the two prompts and altering the distance between these two prompts will affect the model's performance.}
\end{figure*}

\begin{table}[!tb]
\caption{\label{CIFAR10-to-CIFAR10C gradual} \textbf{Gradually changing CIFAR10-to-CIFAR10C results.} The severity level changes gradually between the lowest and the highest. The corruption type changes when the severity is the lowest. Results are the mean over ten diverse corruption type sequences. Our method achieves the best performance.}
\centering
\setlength\tabcolsep{3pt}
\begin{tabular}{lccccc}
\toprule
Error(\%)&Source&BN Adapt&TENT&CoTTA&Ours\\\midrule
CIFAR10C &24.8&13.7&30.7&10.4&6.1\\ \bottomrule
\end{tabular}
\vspace{-0.2cm}
\end{table}

\begin{table}[!tb]
\caption{\label{ImageNet-to-ImageNet-c} \textbf{Average error of standard ImageNet-to-ImageNet-C experiments over 10 diverse corruption sequences (severity level 5).} Our method is 11.5\% higher than the state-of-the-art method.}
\label{imagenetc}
\centering
\setlength\tabcolsep{0.4pt}
\begin{tabular}{lcccccc}
\toprule
Error(\%)&Source&BN Adapt&Test Aug&TENT&CoTTA&Ours\\\midrule
ImageNet-C&82.4&72.1&71.4&66.5&63.0&51.5\\ \bottomrule
\end{tabular}
\vspace{-0.2cm}
\end{table}

\noindent\textbf{ImageNet-to-ImageNet-C}
In order to prove the effectiveness of our proposed method on a more extensive dataset, we conduct experiments on ImagNet-to-ImageNet-C. Table \ref{imagenetc} shows our method achieves the best performance, which is 11.5\% higher than the state-of-the-art method.

\subsection{Performance on real domains shifts}

\begin{table}[!tb]
\caption{\textbf{Ablation: Contribution of our proposed DAP and DSP. 
}}
\label{ablationDAP}
\centering
\setlength\tabcolsep{3pt}
\renewcommand\arraystretch{1}
\begin{tabular}{cccccc}
\toprule
\# & \makecell*[c]{DSP} & \makecell*[c]{DAP} & CIFAR10C & CIFAR100C & ImageNet-C\\\midrule
0 &  & &16.2&32.5& 63.0 \\ 
1 & \checkmark & & 14.9 & 17.8& 52.4\\
2 &  & \checkmark & 15.2 & 19.5& 53.2\\
3 & \checkmark & \checkmark & 13.9 & 16.3& 51.5 \\ \bottomrule
\end{tabular}
\vspace{-0.2cm}
\end{table}


\noindent\textbf{Domain generalization benchmark dataset VLCS.} Compared with CIFAR10-C and ImagenNet-C, there is a real-world and more significant distribution gap between domains in the VLCS task. In addition, the model needs to adapt to the target domains at multiple rounds, making the VLCS task more challenging. According to Table \ref{dg benchmark}, compared with other methods, our method has a lower error, i.e., when VOC2007 is the source domain, and the other three are the target domain. The mean error is 13\% lower than CoTTA. Moreover, the performance in each round will gradually improve, i.e., when CalTech-101 is the source domain, the error on LabelMe in the three rounds will continue to decrease from 67.4\% to 66.1\%. 
Besides, there exists a significant gain (15.9\%) in VOC. We conclude the reason as prompts aim to exploit the knowledge of the source model and adapt to new target domains. Since VOC has richer data and more balanced classes, the source model trained on it can reflect richer and more reliable knowledge, which lays a better foundation for prompt mechanisms. Then for incoming target domains, the prompt mechanism can outperform model renewal approaches in mitigating error accumulation and catastrophic forgetting pains. In contrast, the other two datasets have imbalanced categories(i.e., bird and dog out of the five classes in SUN are only with $\sim$ 2\% images).
These results show our proposed method's robust anti-forgetting performance when the distribution shifts are large.

\subsection{Analysis}
\noindent\textbf{How do the prompts' size and location affect the model's performance?} According to Figure \ref{fig:4}, changing prompt sizes shows a small variance (within 1 $\sim$ 2\%), demonstrating the performance is not sensitive to it. \cite{MenglinJia2022VisualPT} also shows experimentally that prompts with smaller sizes can perform similarly to those with larger sizes. This applies in the patch attack area \cite{AbhijithSharma2022AdversarialPA}, too. As for prompt location, random location performs the best because the location variance of key semantic information across images can be better averaged by random position selection. The curves in Figure \ref{fig:4} appear to be roughly symmetrical. This means the performance of the two prompts differs when simply interchanging the locations, indicating that the domain attributes reflected in different regions on the same image are different. Overall, a random location is a simple but effective choice.

\noindent\textbf{What are the contributions of DAP and DSP, respectively?} We validate the contributions of the proposed DSP and DAP (Table \ref{ablationDAP}). Using DAP and DSP alone will improve the model's performance, and the performance improvement of DSP alone will be more obvious. This is because DSP has targeted domain-specific knowledge, but there is no significant difference between the two compared with DAP alone. The role of domain-agnostic knowledge of DAP and the role of domain-specific knowledge of DSP complement each other in the face of chatting target domains. 

\noindent\textbf{Does using prompt partly destroy the information of the original image?} On the one hand, visual domain prompts are not a mask but a summation relationship with the original image. Therefore, if there is valuable information in a region, the model can learn this and make the value of visual domain prompts equal to 0. On the other hand, in the ablation experiments, we demonstrate that the best performance of the model can be achieved when the location of the visual domain prompts is random. To some extent, the random position can alleviate the coverage of essential information regions. It's interesting to try to add a learnable position setting to explore this question further.

\section{Conclusion}
In this paper, to tackle the error accumulation problem of CTTA, we first introduce a concept of visual domain prompts that are 1) small image tokens and 2) dynamically added upon the input image to shift them from the changing target domains to the regular domain. We then propose a new CTTA framework, Continual Test-time Adaptation via Domain Prompt (CTAP), that consists of the visual domain prompt updating module and the Homeostasis-based adapting strategy. We solve this problem from the perspective of changing the input images by the lightweight domain prompt tokens. Extensive experiments on multiple benchmark datasets demonstrate our method achieves SOTA performance at a relatively small cost.

\section{Acknowledgements}
This research was supported in part by the Foundation of Shenzhen Science and Technology Innovation Committee (JCYJ20180507181527806). We thank Qiuchuan Liang for doing some data processing work. 
\bibliography{aaai23}



\end{document}